# Greedy Block Coordinate Descent for Large Scale Gaussian Process Regression

**Liefeng Bo**[1]
[1]Toyota Technological Institute at Chicago
Chicago, IL 60637
blf0218@tti-c.org

**Cristian Sminchisescu**[2,1]
[2]University of Bonn
Bonn, 53115, Germany
sminchisescu.ins.uni-bonn.de

## Abstract

We propose a variable decomposition algorithm–greedy block coordinate descent (GBCD)–in order to make *dense* Gaussian process regression practical for large scale problems. GBCD breaks a large scale optimization into a series of small sub-problems. The challenge in variable decomposition algorithms is the identification of a sub-problem (the active set of variables) that yields the largest improvement. We analyze the limitations of existing methods and cast the active set selection into a zero-norm constrained optimization problem that we solve using greedy methods. By directly estimating the decrease in the objective function, we obtain not only efficient approximate solutions for GBCD, but we are also able to demonstrate that the method is globally convergent. Empirical comparisons against competing dense methods like Conjugate Gradient or SMO show that GBCD is an order of magnitude faster. Comparisons against sparse GP methods show that GBCD is both accurate and capable of handling datasets of 100,000 samples or more.

## 1 Introduction

Solving linear systems is frequently encountered in least squares kernel methods. A relevant example is Gaussian process regression (GPR) with Gaussian noise (Williams & Rasmussen, 1996; Rasmussen & Williams, 2006), a method that has become increasingly popular in the field of machine learning. Given a set of training samples $\{\mathbf{x}_i\}_{i=1}^n$ along with the corresponding targets $\{y_i\}_{i=1}^n$, the predictive mean and variance of the estimator can be computed in closed form

$$m(\mathbf{x}_*) = \mathbf{k}_*^\top \left(\mathbf{K} + \sigma^2 \mathbf{I}\right)^{-1} \mathbf{y} \qquad (1)$$

$$v(\mathbf{x}_*) = k(\mathbf{x}_*, \mathbf{x}_*) + \sigma^2 - \mathbf{k}_*^\top \left(\mathbf{K} + \sigma^2 \mathbf{I}\right)^{-1} \mathbf{k}_* \qquad (2)$$

where $k(\mathbf{x}_i, \mathbf{x}_j)$ is a positive definite kernel function, $\mathbf{K}$ is a $n \times n$ matrix with $\mathbf{K}_{ij} = k(\mathbf{x}_i, \mathbf{x}_j)$, $\mathbf{k}_*$ is a $n \times 1$ vector with the $i$-th component being $\mathbf{k}_* = [k(\mathbf{x}_*, \mathbf{x}_i)]$, and $\sigma^2$ is noise variance. GPR requires an $n \times n$ matrix inversion with $O(n^3)$ training cost and $O(n^2)$ requirements for memory storage, which is non-trivial since the kernel matrix $\mathbf{K}$ can not be fitted into memory for large datasets (this is known as the out-of-memory case). Many standard linear system solvers, such as Cholesky factorization, implicitly assume the storage of $\mathbf{K}$ is possible, which limits their applicability in this case. Alternatively, one can recompute $\mathbf{K}$ on the fly, but this becomes computationally prohibitive if $\mathbf{K}$ is frequently needed.

Instead of directly inverting the covariance matrix, one can use iterative methods such as the conjugate gradient algorithm (CG) in order to train GPR. This starts with an initial guess and modifies the current solution iteratively until a given stopping condition is satisfied. CG has speed of convergence guarantees in that it takes no more than $n$ steps to reach the exact solution. In practice, the actual number of iterations necessary for a given precision is often much smaller than $n$, hence CG is usually faster than direct matrix inversion. Unfortunately, CG is again not well adapted for GPR in the out-of-memory case because expensive covariance matrix re-evaluations are necessary inbetween iterations.

Another way to attack the out-of-memory case is the block coordinate descent (BCD) method (Bertsekas, 1999), known as the block Gauss-Seidel for linear systems (Ortega & Rheinboldt, 1970). At each iteration, block coordinate descent splits variables into two subsets, the set of the active variables and the set of inactive ones, then minimizes the objective function along active dimensions while inactive variables are fixed at current values. Block coordinate descent is attractive since only the kernel matrix indexed by the active set needs to be (re)evaluated, thus significantly reducing the cost per iteration.

The efficiency of block coordinate descent methods depends strongly on the active set selection. A frequently used method is to select the active set in cyclical order.

Beatson et al. (2000) applied the cyclical block coordinate descent for solving radial basis function interpolation equations and Li et al. (2007) adopted it to train regularized least squares classifiers. The potential weakness of cyclical block coordinate descent is that it does not exploit any information about the objective function in the process of selecting the active set. The gradient-based active set selection is an alternative method that maximizes the gradient norm (Zoutendijk, 1970; Joachims, 1999). However, the gradient-based active set selection still does not directly relate the variable selection and the decrease in the objective function, leading to non-optimal updates.

In this paper, we propose a greedy block coordinate descent (GBCD) method in order to improve the selection of sub-problems solved during optimization. We show that active set selection can be cast as a zero-norm constrained optimization problem. While the exact solution requires combinatorial search, we show that greedy algorithms can be used to obtain approximate solutions with $O(m^3)$ cost, where $m$ is the size of sub-problems. This maintains the same computational complexity as gradient-based active set selection, per iteration, yet it achieves significant speed-ups. Differently from existing active set methods that select all active variables simultaneously, our method constructs the active set incrementally. Hence, it can build upon the current optimization context in order to implicitly avoid redundancy in previously selective active variables. This avoids inefficiencies due to the inclusion of high correlated variables in the active set. Our experiments show that GBCD compares favorably with existing methods, both dense and sparse (and against the constraints expected in each case, i.e. accuracy vs. speed). We do not only present experiments on standard machine learning benchmarks but also on real-world large scale computer vision applications like the reconstruction of three-dimensional human motions like walking, running, gestures or boxing from image sequences acquired with a single video camera, as available in the HumanEva dataset (Sigal & Black, 2006).

### 1.1 Related Work

**Sequential minimal optimization:** Decomposition algorithms (Platt, 1999; Keerthi et al., 2001) have been widely used in the field of support vector machines (SVMs). Sequential minimal optimization (SMO) (Platt, 1999) is the most prominent method in which the working set only includes two variables, hence the sub-problems can be solved analytically. The most popular way to select the active set for SMO is the maximal violating pair method, first proposed by (Keerthi et al., 2001) and used in many SVMs packages such as LIBSVM 2.71 (Chang & Lin, 2001). Keerthi and Shevade (2003) extended SMO to least square support vector machines, a problem very similar with the one we consider here, hence SMO also can be adopted for GPR. SMO differs from our method in the way the active variables are chosen and in the size of sub-problems.

**Fast multipole methods:** The most time-consuming step in CG is the multiplication of a kernel matrix with a vector. This can be sped up by fast multipole methods and KD-Trees. In particular, Yang et al. (2005) and Shen et al. (2006) have applied this method to Gaussian process regression. While the methods appear to be effective on low-dimensional problems, they have not been demonstrated yet for high dimensional problems. The quality of Hermite or Taylor approximations used may also degrade exponentially as a function of the dataset dimensionality.

**Sparse GPR models:** Recently, there has been substantial research on deriving sparse approximations to the full GPR (Smola & Bartlett, 2001; Csató & Opper, 2002; Lawrence et al., 2003; Seeger et al., 2003; Keerthi et al., 2006; Keerthi & Chu, 2006) that reduce the training and storage cost. The methods select a representative subset of regressors, thus dropping the training complexity to $O(np^2)$, where $p$ is the size of the subset. Since $p \ll n$ in most cases, sparse approximations achieve substantial speedups relative to the full GPR. Though very appealing, sparse approximations are not always the best choice in applications that require high accuracy. Indeed, recent research (Rasmussen & Candela, 2005; Quiñonero-Candela & Rasmussen, 2005) confirms that sparse GP can sometimes lead to unreasonable predictive distributions. Our experiments also show that sparse GP methods can sometimes be significant less accurate compared to the full GPR model.

## 2 Greedy Block Coordinate Descent

It is well-known that the solution to the positive definite linear system (1) and (2) can be obtained from a quadratic optimization problem

$$\min_{\alpha} \left[ f(\alpha) = \frac{1}{2} \alpha^\top \bar{\mathbf{K}} \alpha - \mathbf{y}^\top \alpha \right] \quad (3)$$

where $\bar{\mathbf{K}} = \mathbf{K} + \sigma^2 \mathbf{I}$.

Let $\mathbf{g}^k = \bar{\mathbf{K}} \alpha^k - \mathbf{y}$ be the current gradient vector. A gradient-based method selects the active set $B$ by solving the following optimization problem

$$\max_{|B|=m, B \subseteq \{1,\cdots,n\}} |\mathbf{g}_B^k| \quad (4)$$

where $m$ is the size of active set. A key observation is that the gradient-based active set selection is not directly related with the decrease in the objective function, possibly leading to non-optimal improvement. In contrast, we integrate the active set selection into the solution to sub-problems. At each iteration of decomposition, it is desirable to decrease the objective function as much as possible for a given size of the active set. The zero-norm formulation of this prob-

**Algorithm 1** Greedy Block Coordinate Descent

1. Set $k = 0$, $\alpha^k = \mathbf{0}$, and $\mathbf{g}^k = -\mathbf{y}$

2. If $\alpha^k$ is an optimal solution of (3), stop; otherwise go to step 3

3. Solve (6) using greedy algorithm and obtain $\Delta\alpha^{opt}$ and the active set $B$

4. Set $\alpha_B^{k+1} = \alpha_B^k + \Delta\alpha_B^{opt}$, $\alpha_N^{k+1} = \alpha_N^k$, and $\mathbf{g}^{k+1} = \mathbf{g}^k + \mathbf{K}_B \Delta\alpha_B^{opt}$ where $\mathbf{K}_B$ is the sub-matrix made of the columns indexed by $B$ and $\alpha_B^k$ is the sub-vector indexed by $B$. Let $k = k + 1$ and go back to step 2

lem can be written as

$$\min_{\Delta\alpha} \left[ \frac{1}{2}(\alpha^k + \Delta\alpha)^\top \bar{\mathbf{K}}(\alpha^k + \Delta\alpha) - \mathbf{y}^\top(\alpha^k + \Delta\alpha) \right] \quad (5)$$
$$s.t. \quad \|\Delta\alpha\|_0 = m$$

where $\|\cdot\|_0$ is the zero-norm, counting the nonzero entries of a vector, and $m$ is the number of nonzero entries, i.e. the size of the active set. Eliminating the constant term in (5), we obtain

$$\min_{\Delta\alpha} \left[ \frac{1}{2}\Delta\alpha^\top \bar{\mathbf{K}} \Delta\alpha + (\mathbf{g}^k)^\top \Delta\alpha \right] \quad (6)$$
$$s.t. \quad \|\Delta\alpha\|_0 = m$$

where $\mathbf{g}^k = \mathbf{g}(\alpha^k) = \bar{\mathbf{K}}\alpha^k - \mathbf{y}$. There are several difficulties in solving (6). First, the constraint is not differentiable, so gradient descent algorithms can not be used. Second, the optimizers can get trapped in a shallow local minimum because the cost (6) is not necessary convex. An exhaustive search over all possible choices ($\|\Delta\alpha\|_0 = m$) is expensive as the number of possible combinations $\binom{n}{m}$ is usually too large to enumerate even on current computers. One option is to use more sophisticated search algorithms such as branch-and-bound in order to decrease the cost of exhaustive search. But, branch-and-bound is still too expensive for large $n$, and it seems unwise to spend too much work on sub-problems anyway. In fact, although the approach can decrease the objective faster and may take fewer iterations to converge, the overall training time may not be reduced since each iteration is expensive.

A more practical view is to decrease the objective function as much as possible with little extra work. In this paper, we compute an approximate solution to (6) using a greedy algorithm. This gives a balanced tradeoff between the decrease in objective function and the computational cost of each iteration, and converges fast in our experiments. We refer to this novel decomposition algorithm as Greedy Block Coordinate Descent (GBCD), and its steps are described in the table associated with **Algorithm 1** (Greedy Block Coordinate Descent).

### 2.1 Greedy Approximation

Unlike cyclical and gradient-based active set methods that select all variables simultaneously, our greedy algorithm selects the active variables incrementally. Starting with an empty active set, our greedy optimizer selects one active variable at a time, so that the objective function is decreased. The selection process stops when the number of active variables reaches a predefined value $m$.

Let $B = \emptyset$ and $N = \{1, \cdots, n\}$. How do we select an active variable from $N$? A natural idea is to optimize the objective with respect to $\Delta\alpha_B$ and $\Delta\alpha_i$ for each $i \in N$ and select the variable giving the largest decrease. This process leads to a two-layer optimization problem

$$s = \arg\min_{i \in N} \left[ \min_{\Delta\alpha_B, \Delta\alpha_i} \frac{1}{2} \begin{bmatrix} \Delta\alpha_B \\ \Delta\alpha_i \end{bmatrix}^\top \begin{bmatrix} \bar{\mathbf{K}}_{BB} & \bar{\mathbf{K}}_{Bi} \\ \bar{\mathbf{K}}_{iB} & \bar{\mathbf{K}}_{ii} \end{bmatrix} \begin{bmatrix} \Delta\alpha_B \\ \Delta\alpha_i \end{bmatrix} \right.$$
$$\left. + \begin{bmatrix} \mathbf{g}_B^k \\ g_i^k \end{bmatrix}^\top \begin{bmatrix} \Delta\alpha_B \\ \Delta\alpha_i \end{bmatrix} \right] \quad (7)$$

where $\bar{\mathbf{K}}_{Bi}$ is the sub-matrix made of rows indexed by $B$ and the column indexed by $i$. This active set selection method, called prefitting, has appeared in the (prefitting) version of kernel matching pursuit (KMP) (Vincent & Bengio, 2002) and the sparse greedy Gaussian process (Smola & Bartlett, 2001). However, prefitting needs to solve $m+1$ dimensional optimizations $|N|$ times, which obviously has a higher cost than optimizing the sub-problem.

A cheaper method is to fix $\Delta\alpha_B^t$ and optimize (7) only with respect to $\Delta\alpha_i$. This can be expressed as a two-layer optimization problem

$$s = \arg\min_{i \in N} \left[ h_i = \min_{\Delta\alpha_i} \frac{1}{2} \begin{bmatrix} \Delta\alpha_B^t \\ \Delta\alpha_i \end{bmatrix}^\top \begin{bmatrix} \bar{\mathbf{K}}_{BB} & \bar{\mathbf{K}}_{Bj} \\ \bar{\mathbf{K}}_{iB} & \bar{\mathbf{K}}_{ii} \end{bmatrix} \begin{bmatrix} \Delta\alpha_B^t \\ \Delta\alpha_i \end{bmatrix} \right.$$
$$\left. + \begin{bmatrix} \mathbf{g}_B^k \\ g_i^k \end{bmatrix}^\top \begin{bmatrix} \Delta\alpha_B^t \\ \Delta\alpha_i \end{bmatrix} \right] \quad (8)$$

Eliminating the constant in (8), we get

$$h_i = \min_{\Delta\alpha_i} \left[ \frac{1}{2}\left( k(\mathbf{x}_i, \mathbf{x}_i) + \sigma^2 \right) \Delta\alpha_i^2 + e_i^t \Delta\alpha_i \right], \quad i \in N \quad (9)$$

where $e_i^t = \bar{\mathbf{K}}_{iB} \Delta\alpha_B^t + g_i^k$. This is a one dimensional quadratic programming problem and can be solved analytically. Thus, we can determine the new active variable by the formula

$$s = \arg\min_{i \in N} \left[ h_i = \frac{-(e_i^t)^2}{2(k(\mathbf{x}_i, \mathbf{x}_i) + \sigma^2)} \right] \quad (10)$$

In solving the sub-problems, the inversion of the covariance matrix is the computational bottleneck. In the $(t+1)$-th iteration, the inversion of the covariance indexed by the current active set can be written as

$$\mathbf{R}^{t+1} = \begin{bmatrix} \bar{\mathbf{K}}_{BB} & \bar{\mathbf{k}}_s \\ \mathbf{k}_s^\top & \bar{\mathbf{K}}_{ss} \end{bmatrix}^{-1} \quad (11)$$

where $\bar{\mathbf{k}}_s = [\bar{\mathbf{K}}_{b_1 s}, \bar{\mathbf{K}}_{b_2 s}, \cdots, \bar{\mathbf{K}}_{b_r s}]$. Applying the Woodbury inversion identity (Stoer & Bulirsch, 1993) to (11), we get

**Algorithm 2** Greedy Approximation with Random Subset

1. Set $t=0$, $\Delta\alpha^t = \mathbf{0}$, $\mathbf{e}^t = \mathbf{g}^k$, $B = \emptyset$, and $O = N = \{1, \cdots, n\}$

2. If $t = m$, stop; otherwise go to step 3

3. $s = \arg\min_{i \in O} \left[ \dfrac{-(e_i^t)^2}{2(k(\mathbf{x}_i, \mathbf{x}_i) + \sigma^2)} \right]$

4. If t=0, $\mathbf{R}^{t+1} = \dfrac{1}{k(\mathbf{x}_i, \mathbf{x}_i) + \sigma^2}$ and $\Delta\alpha_s = \dfrac{-e_s}{k(\mathbf{x}_i, \mathbf{x}_i) + \sigma^2}$, otherwise compute $\mathbf{R}^{t+1}$, $\Delta\alpha_B^{t+1}$, and $\Delta\alpha_s^{t+1}$ according to (12) and (13)

5. Set $B = B + \{s\}$, $N = N - \{s\}$

6. Randomly choose a subset $O$ of size $\kappa$ from $N$. Let $\mathbf{e}_O^{t+1} = \bar{\mathbf{K}}_{OB} \Delta\alpha_B^{t+1} + \mathbf{g}^k$, $t = t + 1$, and go to step 2

an update formula

$$\mathbf{R}^{t+1} = \begin{bmatrix} \mathbf{R}^t & \mathbf{0} \\ \mathbf{0}^\top & 0 \end{bmatrix} + \eta \begin{bmatrix} \boldsymbol{\beta} \\ -1 \end{bmatrix} \begin{bmatrix} \boldsymbol{\beta}^\top & -1 \end{bmatrix} \quad (12)$$

where $\boldsymbol{\beta} = \mathbf{R}^t \bar{\mathbf{k}}_s$, $\eta = (\bar{K}_{ss} - \bar{\mathbf{k}}_s^\top \boldsymbol{\beta})^{-1}$. Combining (12) and $\Delta\alpha_B^t = -\mathbf{R}^t \mathbf{g}_B^k$, we obtain the update formula for $\Delta\alpha^{t+1}$

$$\begin{bmatrix} \Delta\alpha_B^{t+1} \\ \Delta\alpha_s^{t+1} \end{bmatrix} = -\mathbf{R}^{t+1} \begin{bmatrix} \mathbf{g}_B^k \\ g_s^k \end{bmatrix} = \begin{bmatrix} \Delta\alpha_B^t \\ 0 \end{bmatrix} - \eta \left( \boldsymbol{\beta}^\top \mathbf{g}_B^k - g_s^k \right) \begin{bmatrix} \boldsymbol{\beta} \\ -1 \end{bmatrix} \quad (13)$$

(12) and (13) indicate that we can efficiently update $\mathbf{R}^{t+1}$ and $\Delta\alpha^{t+1}$ at a cost of $O(t^2)$ without explicitly computing the inverse matrix. The update formula (12) is numerical stable since the regularization term $\sigma^2 \mathbf{I}$ greatly improves the condition number of the matrix $\mathbf{K}$. One can improve the numerical stability further, by using Cholesky decomposition (Stoer & Bulirsch, 1993). Integrating the active set selection and the solution of the sub-problems, we obtain the greedy approximation algorithm shown in the table associated with **Algorithm 2** (Greedy Approximation with Random Subset).

The greedy approximation involves three operations: 1) computing the new column of the covariance which is $O(cn)$, where $c$ is the cost of evaluating the kernel function one time, 2) updating $\mathbf{e}^t$ which is $O(nm)$, and 3) selecting the new active variable according to (10) which is $O(n)$. The cost of solving sub-problems is dominated by updating the inverse of the covariance sub-matrix which is $O(m^2)$. Adding up, the cost of choosing $m$ active variables gives the overall complexity $O(cnm + nm^2 + m^3)$.

For large $m$ values, updating $\mathbf{e}^t$ dominates the cost of greedy optimizer, which is still more than what we want to accept. This cost can be reduced to $O(\kappa m^2)$ by only considering the random subset $O$ of $N$ with size $\kappa$ and selecting the active variables only from $O$ rather than exhaustively searching the full set $N$. Note that the number of covariance function evaluations at each iteration is still $O(cnm)$ because step 4 of GBCD requires computing a covariance matrix indexed by the active set.

## 2.2 Convergence

**Theorem 1**: Let

$$\lambda_{min} = \min_{B \subseteq \{1, \cdots, n\} : |B| = m} \left[ \min eig\left( \bar{\mathbf{K}}_{BB} \right) \right] \quad (14)$$

where $\min[eig(\bar{\mathbf{K}}_{BB})]$ denotes the smallest eigenvalue of the matrix $\bar{\mathbf{K}}_{BB}$. The following inequation holds

$$f\left(\boldsymbol{\alpha}^{k+1}\right) - f\left(\boldsymbol{\alpha}^k\right) \leq -\frac{1}{2} \lambda_{min} \left\| \boldsymbol{\alpha}^{k+1} - \boldsymbol{\alpha}^k \right\|_2^2 \quad (15)$$

Theorem 1 indicates that a random chosen active set can give the objective function (3) sufficient decrease if the covariance function is positive definite.

**Theorem 2**: $\{\alpha^k\}$ converges to the global minimum of (3).

Proofs to Theorems 1 and 2 are given in the Appendix.

## 2.3 Intuitions on Active Set Selection

This section analyses the efficiency of decomposition algorithms and motivates why the greedy active set selection we propose works better than competing methods. We attribute the efficiency of our decomposition to its gains over *gradient correlation*: if inputs $\mathbf{x}_i$ and $\mathbf{x}_j$ are neighbors, their gradients $g_i$ and $g_i$ are correlated with high probability. In contrast, our method selects variables with *uncorrelated* gradients. We validate this conjecture further by means of a quantitative experiment, where we randomly choose an input $\mathbf{x}_i$ and find its 50 neighbors based on the Euclidean distance between $\mathbf{x}_i$ and the other inputs. We then track the optimization process of GBCD and record the gradient of $\mathbf{x}_i$ and its neighbors across 50 successive iterations. Finally, we compute correlation coefficients for the gradient of $\mathbf{x}_i$ and its neighbors using $\dfrac{\text{cov}(g_i, g_j)}{\text{std}(g_i)\text{std}(g_j)}$. We repeat the process over 100 inputs randomly chosen from training set. Counting the frequency of correlation coefficients falling in different intervals, we obtain the histograms in fig. 1. One can see that the gradients of $\mathbf{x}_i$ and its neighbors are non-negligibly correlated.

Intuitively, one should select the active set in a way that makes the gradients as uncorrelated as possible, because this brings in diversity from variables outside the active set. This explains our experimental findings: the cyclical active set selection works better than the gradient-based one. The latter works by selecting variables that maximize the gradient infinite norm, thus it tends to select many correlated variables, which prevents rapid progress during optimization. The cyclical method avoids this on average, because it is random; similarly does SMO, which selects the maximal violating pair as the active set—this is usually uncorrelated with high probability. Our greedy method selects the active set incrementally, hence it exploits information in the previously chosen active variables and avoids the selection of highly correlated ones. The gradients corresponding to variables that are highly correlated with ones previously selected are usually small–in contrast the other greedy methods will likely select variables with simply large gradients, see (10).

Table 1: Number of training and test samples, and size of attribute vector from benchmark datasets. No matter the algorithm, the samples we test on are never used for validation or training.

| Problem | Training | Test | Attribute |
|---|---|---|---|
| Calhouse | 18000 | 2640 | 8 |
| Outaouais | 20000 | 9000 | 37 |
| Kin40k | 30000 | 10000 | 8 |
| Sarcos | 44484 | 4449 | 21 |
| Friedman1 | 100000 | 5000 | 10 |

## 3 Experiments

In this section, we empirically study the behavior of GBCD on several benchmark datasets and compare it to existing methods, both dense and sparse. The benchmarks are presented in table 1. Kin40k are available from Torgo's homepage[1], Outaouais is from the *Evaluating Predictive Uncertainty Challenge*[2], Friedman1 is from the author, Friedman (1991), Sarcos from the book *Gaussian processes for machine learning*[3]. Gaussian noise with unit standard deviation is added to the training samples. For all datasets, each attribute of the training inputs and outputs are linearly scaled to zero mean and unit variance and the same transformation is used for the test set. Unless otherwise specified, we report the normalized root mean squared error (RMSE) on the test set given by

$$RMSE = \sqrt{\frac{1}{t}\sum_{i=1}^{t}\frac{(y'_i - m_i)^2}{var(\mathbf{y})}} \quad (16)$$

where $y'_i$ is the output of test samples, $m_i$ is the predictive mean and $var(\mathbf{y})$ is the variance of training samples. For algorithms that require random numbers, RMSE is averaged over 10 trials.

All algorithms are implemented in VC++ 6.0 and run on a PC with 3.6 GHz P4 processors, 2 GB memory, and Windows XP. Greedy approximations based on random search are used. The size of the random subset $\kappa$ is set to 60. (Smola and Schölkopf (2000) have also found that the random set of size 59 achieved good performance). Optimization is assumed converged when the infinite norm of gradient is within $10^{-4}$ tolerance. The size of sub-problems is set to 500 for BCDC, BCDG and GBCD.

---
[1] http://www.liacc.up.pt/~ltorgo/Regression/DataSets.html
[2] http://predict.kyb.tuebingen.mpg.de
[3] http://www.gaussianprocess.org/gpml/data/

The squared exponential function $k(\mathbf{x}_i, \mathbf{x}_j) = \exp(-\sum_{l=1}^{d} \gamma_l (\mathbf{x}_i^l - \mathbf{x}_j^l)^2)$ is used to construct the kernel matrix. The optimal values for hyperparameters $(\gamma_1, \cdots, \gamma_d, \sigma^2)$, are determined, for each model, by maximizing the marginal likelihood on a tractable subset of 2000 samples randomly sampled from the training set. BFGS is used to optimize the hyperparameters. The termination tolerance on the marginal likelihood and on the hyperparameters is set to $10^{-4}$.

To compute the predictive mean of GPR, we only need to solve one linear system whereas in order to compute the predictive variance, we need to solve a linear system per sample. In the sequel, *training time* is measured as time required to solve the linear system $\mathbf{K} + \sigma^2 \mathbf{I} = \mathbf{y}$; *mean time* is the time necessary to compute the predictive mean; *variance time* is the time needed to compute the predictive variance, which requires solving one linear system $\mathbf{K} + \sigma^2 \mathbf{I} = \mathbf{k}_*$. If the computation variance is required, this will dominate test time.

### 3.1 Comparisons with Other Iterative Methods

This section compares CG, SMO, Cholesky factorization, BCDC, BCDG and GBCD. BCDC selects an active set using the cyclic order and BCDG selects an active set using gradient information. We only use a subset of 10,000 samples randomly chosen from training samples for all algorithms, which allows us store the full covariance matrix and solve GPR using Cholesky factorization. Cholesky factorization serves as baseline, in order to know whether the solutions obtained by the iterative algorithms we test are close to the exact one. We do not include the gradient-based active set selection method because the cyclic method outperforms it in all of our experiments. For CG, SMO, BCDC, BCDG and GBCD, we re-calculate the covariance matrix whenever needed in order to simulate the out-of-memory case.

Table 2 gives the training time and RMSE of four algorithms. GBCD is significantly faster than CG, SMO, BCDC and BCDG on all the datasets. BCDG is significantly slower than other methods because of selection of many correlated variables. In particular, for Sarcos, GBCD is 10 times faster than BCDC, 20 times faster than SMO and 59 times faster than CG. CG, SMO, BCDC, BCDG and GBCD have RMSE accuracy that is similar to the result of Cholesky factorization up to 3 significant digits. This confirms that our iterative algorithms find good solutions for the convergence tolerances used.

Fig. 2 shows the RMSE, gradient infinite norm and objective function on the test set, for all algorithms as function of training time. We do not include the test error of CG and BCDG because these give significantly higher test errors at the initial stages of optimization and affects the scaling / readability of plots even on log-scale. GBCD con-verges significantly faster than other algorithms in terms of both the objective function decrease and the gradient infinite norm. GBCD usually achieves stable test accuracy far before the specified convergence criterion is reached. This indicates that a looser criterion can be used when training time constraints exist.

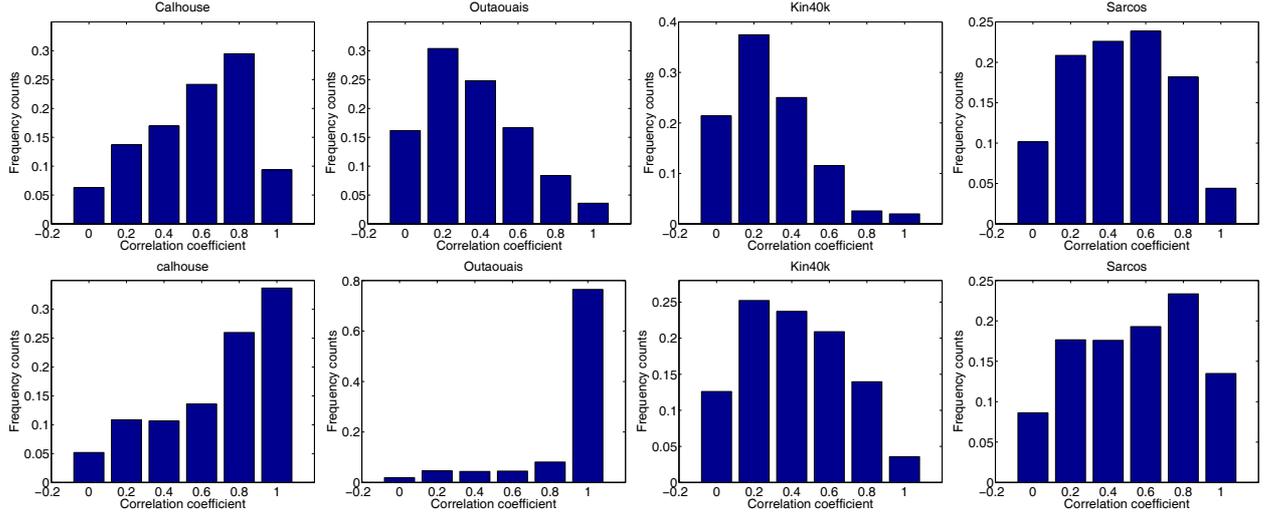

Figure 1: Frequency of correlation coefficients of neighboring points, histogramed. Plots on top row are obtained by solving the linear system $\mathbf{K} + \sigma^2 \mathbf{I} = \mathbf{y}$, the ones on the bottom row are obtained by solving the linear system $\mathbf{K} + \sigma^2 \mathbf{I} = \mathbf{k}_*$. Correlation coefficients for the regime shown on the bottom are higher than the corresponding ones on the top. This indicates that the linear system $\mathbf{K} + \sigma^2 \mathbf{I} = \mathbf{k}_*$ is easier to solve, which is consistent with our experiments, discussed in §3.

Fig. 3 plots the gradient infinite norm and the objective function across iterations. As expected, greedy active set selection decreases the objective significantly faster than active set section based on cyclic ordering, or gradient based active set selection, further confirming our analysis given in §2. 2.

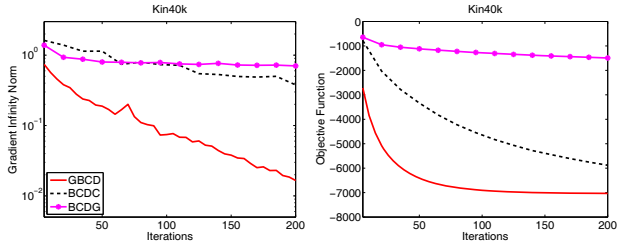

Figure 3: Objective function and infinite gradient norm among iterations. Notice that GBCD gives significantly faster objective decrease than BCDC and BCDG.

Fig. 4 plots training time as function of the size of dataset Friedman1. Notice that the gap between BCDC and GBCD further increases with the size of the dataset, showing that it is impractical to train large scale models using BCDC.

Table 3 gives the averaged variance time of GBCD on each test sample and RMSE of the predictive variance relative to

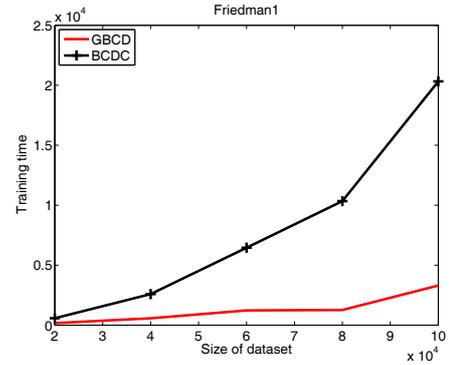

Figure 4: Training time as a function of the dataset size for different models. The training time gap widens as the dataset increases, showing that it is impractical to train BCDC models in this case. BCDC and GBCD are assumed converged when the infinite gradient norm is smaller than $10^{-4}$.

Cholesky factorization

$$\sqrt{\frac{1}{t} \sum_{i=1}^{t} \left( \frac{v'_i - v_i}{v'_i} \right)^2} \qquad (17)$$

where $v'_i$ is the predictive variance of Cholesky, $v_i$ is the predictive variance of GBCD. As one can see, the relative error is 0.02 in the worst case. This indicates that the predictive variance of GBCD is very close to that of Cholesky factorization. We are not able to report the relative RMSE of the predictive variance of the other algorithms relative to that of Cholesky, due to extremely long runtime. Unlike the predictive mean where one only needs to solve a linear system, one needs to solve different linear systems for each test sample in order to obtain the predictive variance.

Table 2: Training time and RMSE on test set of CG, SMO, BCDC, BCDG and GBCD. Chol denotes Cholesky factorization. Optimization is stopped when the infinite norm of the gradient is smaller than $10^{-4}$. '/' show that the values are not available. Bold indicates the lowest training time among all iterative methods. Cholesky factorization stores the full covariance matrix while other method don't.

|  | Training Time (second) | | | | | | Root Mean Squared Error | | | | | |
|---|---|---|---|---|---|---|---|---|---|---|---|---|
|  | Chol | CG | SMO | BCDC | BCDG | GBCD | Chol | CG | SMO | BCDC | BCDG | GBCD |
| Calhouse | 128 | 1958 | 231 | 331 | 25563 | **92** | 0.477 | 0.477 | 0.477 | 0.477 | 0.477 | 0.477 |
| Outaouais | 136 | 24408 | 10320 | 5366 | >10 hours | **1154** | 0.216 | 0.216 | 0.216 | 0.216 | / | 0.216 |
| Kin40k | 130 | 14956 | 5910 | 3677 | >10 hours | **975** | 0.098 | 0.098 | 0.098 | 0.098 | / | 0.098 |
| Sarcos | 132 | 12880 | 4295 | 2172 | >10 hours | **217** | 0.128 | 0.128 | 0.128 | 0.128 | / | 0.128 |
| Friedman1 | 126 | 4732 | 32899 | 122 | >10 hours | **106** | 0.017 | 0.017 | 0.017 | 0.017 | / | 0.017 |

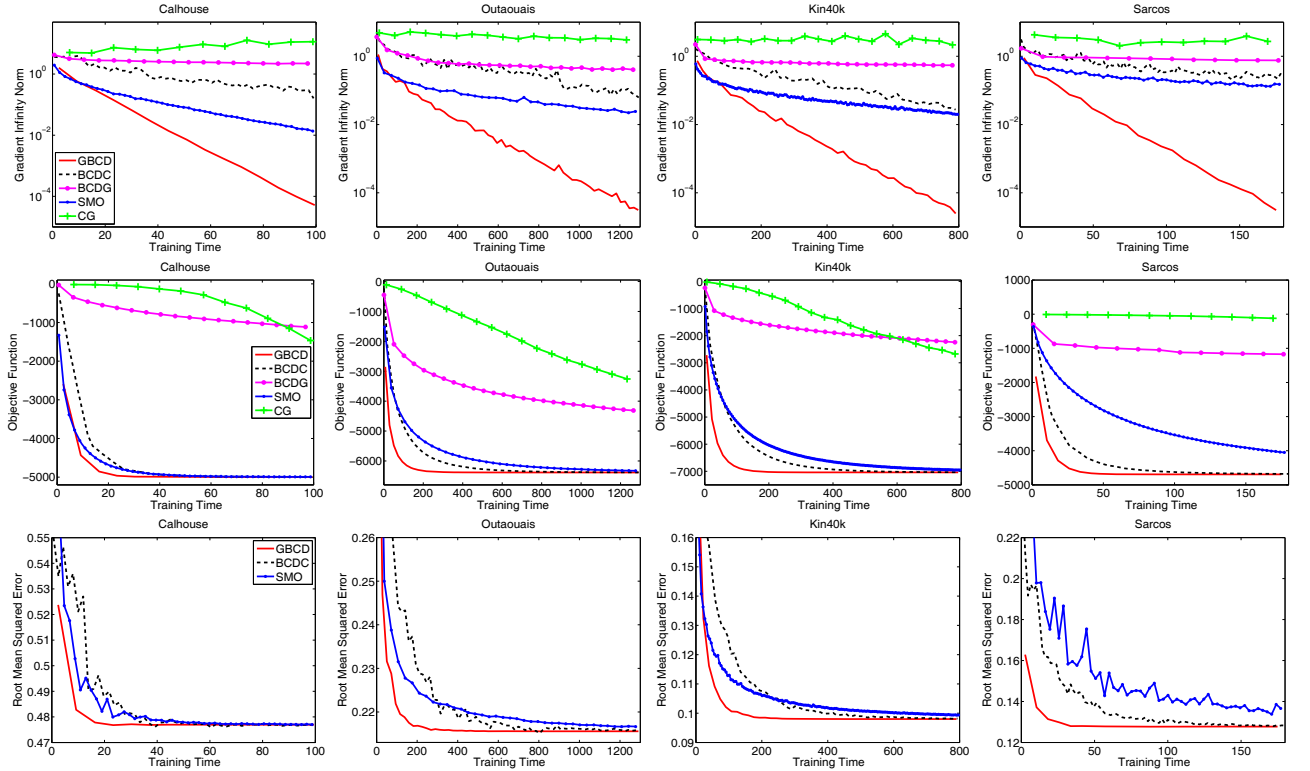

Figure 2: Infinite norm of the gradient, objective function and test error of CG, SMO, BCDC, BCDG and GBCD function of training time, in seconds. The infinite norm of the GBCD gradient converges to $10^{-4}$ precision earlier than competing methods. A similar behavior is observed for the objective function and the test error of GBCD, which stabilize more rapidly than for the other algorithms we have tested.

### 3.2 Comparisons with Sparse GPR

This section compares GBCD and three sparse GPR models. The first is referred as SD (Subset of the Data), which estimates a GPR predictor based on a subset of size $p$ in the training set. The second is known as SR which only uses a random subset of regressors. The third is MPGP which selects the subset of regressors using matching pursuit. The memory requirement of SR and MPGP is $O(pn)$. The size of the subset, $p$ is set to 2000 for all five datasets.

From Table 4, we see that GBCD always achieves lower RSME than sparse GPR models. For Outaouais, GBCD achieves a 55% improvement relative to the best sparse GPR model. This is not surprising because GBCD computes the predictive mean and variance of the full GPR while sparse GPR only gives an approximation. An obvious limitation of sparse GPR models is that computationally feasible $p$ values my lead to less accurate solutions compared to the full GPR. On the other hand, sparse GPR models are faster in testing than GBCD, hence GBCD is well fitted for for problems that require accuracy whereas sparse GPR models are suitable for applications that re-quire fast testing speed. GBCD can be eventually used also in a sparse GPR setting with very large datasets–in this case even representative (sparse) subsets would be large.

Table 4: Training time, variance time and root mean squared error on the test set for GBCD and sparse GPR models. Time is given in seconds. All methods use the same hyperparameters, see section 3.1 for details. The *variance time* is needed to compute the predictive variance. The sparse GPR model is stopped when the specified number of regressors is reached. Convergence for GBCD is assumed when gradient falls below $10^{-4}$. Lowest test error among all methods shown in boldface.

|  | Training Time (second) | | | | Variance Time/sample (second) | | | | Root Mean Squared Error | | | |
| --- | --- | --- | --- | --- | --- | --- | --- | --- | --- | --- | --- | --- |
|  | SD | SR | MPGP | GBCD | SD | SR | MPGP | GBCD | SD | SR | MPGP | GBCD |
| Calhouse | 3 | 48 | 514 | 221 | 0.02 | 0.02 | 0.02 | 19 | 0.514 | 0.477 | 0.475 | **0.472** |
| Outaouais | 5 | 61 | 627 | 5021 | 0.02 | 0.02 | 0.02 | 12 | 0.434 | 0.326 | 0.221 | **0.122** |
| Kin40k | 3 | 85 | 815 | 4233 | 0.02 | 0.02 | 0.02 | 48 | 0.386 | 0.164 | 0.113 | **0.071** |
| Sarcos | 4 | 142 | 1389 | 1879 | 0.02 | 0.02 | 0.02 | 31 | 0.169 | 0.128 | 0.121 | **0.107** |
| Friedman1 | 3 | 231 | 2721 | 3301 | 0.02 | 0.02 | 0.02 | 75 | 0.031 | 0.015 | 0.012 | **0.009** |

Table 3: Average variance time of GBCD on each test sample and root mean square error of the predictive variance relative to that of Cholesky factorization. Variance time denotes the time of computing the predictive variance. Time is given in seconds. GBCD is stopped when the infinite norm of the gradient is smaller than $10^{-4}$.

|  | Variance Time/sample | Relative RMSE |
| --- | --- | --- |
| Calhouse | 12 | 0.00004 |
| Outaouais | 6 | 0.001 |
| Kin40k | 20 | 0.02 |
| Sarcos | 6 | 0.002 |
| Friedman1 | 4 | 0.001 |

### 3.3 Human Pose Estimation

This section presents our results on the HumanEva-1 dataset (Sigal & Black, 2006), a computer vision database that contains a number of human motion sequences of walking, jogging, throw-catch, gestures, and boxing (the backgrounds are known and fairly uniform, hence silhouettes can be computed). See (Bo et al., 2008; Bo & Sminchisescu, 2008; Sminchisescu et al., 2006) for alternative predictors, image representatons and results reported for this problem. The training set consists of pairs of human (image) silhouette-based descriptors and three-dimensional human poses (obtained using a motion capture system, that delivers synchronized images of people and corresponding information about their 3D pose) represented as 45d vectors of three-dimensional body joint positions (a 3d position per human body joint x 15 joints). All poses are preprocessed by subtracting the root joint location from the other joint centers for every frame. We use datasets corresponding to the same set of human motions, captured from three different cameras: C1, C2, C3. The human silhouettes are represented using histograms of semi-local shape contexts – a descriptor that encodes the spatial arrangement of edges around a given image location. To compute shape context descriptors (HistoSC), contours are extracted from the silhouette image and 400 points are sampled on its edges, both internal and external. The shape context descriptor at each point is computed based on 15 angular bins and 8 radial bins, surrounding it. The SC at each of the 400 sample points are accumulated over images subsampled from the training set (typically every 15) and used to generate a codebook using vector quantization. The codebook has 300 clusters obtained using k-means (hence the descriptor size is 300). The descriptor of a new human silhouette, e.g. for testing, is obtained by extracting shape contexts on the edges and vector-quantizing with respect to the existing codebook.

Table 5: Training time and mean joint error of MPGP and GBCD on HumanEva 1. Time is given in seconds and error given in mm, normalized per three-dimensional human body joint location.

|  | Training Time | | Mean Joint Error | |
| --- | --- | --- | --- | --- |
|  | MPGP | GBCD | MPGP | GBCD |
| HistoSC | 5237 | 3891 | 62.4 | 59.6 |

Training and test sets consist of 26572 and 8757 samples, respectively. The size of the subset of regressors is set to 2000 for MPGP. We report the mean joint error in table 5. As expected, our method is more accurate than sparse GPR models. The point is not that much to re-confirm an intuitive outcome, but to show that it is practical to obtain more accurate results on large datasets, otherwise only approachable using sparse methods.

## 4 Conclusions

We have presented an efficient variable decomposition algorithm that is capable of solving large scale Gaussian Process regression problems. The algorithm, referred to as GBCD converges to the global optimum of the objective function and can accurately handle large data sets of 100,000 training samples or more. We have shown that GBCD offers competitive solutions both in terms of accuracy and in terms of training/test time, and it compares favorably with dense and sparse GP methods on machine learning and computer vision datasets. Although we have focused on Gaussian process regression, GBCD is potentially relevant for the solving kernel-based linear systems arising in other models or applications. The idea of breaking a large scale optimization into a series of smaller subproblems and then selecting an active set by objective-sensitive greedy algorithms is quite general. We hope that this strategy will be useful for other optimization problems.

# Appendix

**Proof of Theorem 1**

From the running process of GBCD, we have

$$f\left(\alpha^{k+1}\right) - f\left(\alpha^k\right) = \frac{1}{2}\left(\Delta\alpha_B^{opt}\right)^\top \bar{\mathbf{K}}_{BB}\Delta\alpha_B^{opt} + \left(\mathbf{g}_B^k\right)^\top \Delta\alpha_B^{opt} \quad (18)$$

Substituting $\bar{\mathbf{K}}_{BB}\Delta\alpha_B^{opt}$ into (18), we have

$$f\left(\alpha^{k+1}\right) - f\left(\alpha^k\right) = -\frac{1}{2}\left(\Delta\alpha_B^{opt}\right)^\top \bar{\mathbf{K}}_{BB}\Delta\alpha_B^{opt} \quad (19)$$

Since $\bar{\mathbf{K}}_{BB}$ is a positive definite matrix, there exists one orthonormal matrix $\mathbf{U}$ such that $\bar{\mathbf{K}}_{BB} = \mathbf{U}\Lambda\mathbf{U}^\top$, where $\Lambda = diag\left(\lambda_1(B), \cdots, \lambda_m(B)\right)$. Thus we have

$$\begin{aligned}\left(\Delta\alpha_B^{opt}\right)^\top \bar{\mathbf{K}}_{BB}\Delta\alpha_B^{opt} &= \left(\Delta\alpha_B^{opt}\right)^\top \mathbf{U}\Lambda\mathbf{U}^\top \Delta\alpha_B^{opt} \\ &= \sum_{i=1}^m \lambda_i(B)\left(\mathbf{U}^\top \Delta\alpha_B^{opt}\right)_i^2 \quad (20) \\ &\geq \lambda_{min}(B)\sum_{i=1}^m \left(\mathbf{U}^\top \Delta\alpha_B^{opt}\right)_i^2\end{aligned}$$

where $\lambda_{min}(B) = \min_{1\leq i \leq m}(\lambda_i(B))$. Since $\mathbf{U}$ is orthonormal, we have

$$\begin{aligned}\lambda_{min}(B)\sum_{i=1}^m \left(\mathbf{U}^\top \Delta\alpha_B^{opt}\right)_i^2 &= \lambda_{min}(B)\left(\Delta\alpha_B^{opt}\right)^\top \mathbf{U}\mathbf{U}^\top \Delta\alpha_B^{opt} \\ &= \lambda_{min}(B)\left(\Delta\alpha_B^{opt}\right)^\top \Delta\alpha_B^{opt} \quad (21) \\ &\geq \lambda_{min}\left(\Delta\alpha_B^{opt}\right)^\top \Delta\alpha_B^{opt}\end{aligned}$$

Substituting (20) into (21), we get

$$\begin{aligned}f\left(\alpha^{k+1}\right) - f\left(\alpha^k\right) &\leq -\frac{1}{2}\lambda_{min}\left(\Delta\alpha_B^{opt}\right)^\top \Delta\alpha_B^{opt} \\ &= -\frac{1}{2}\lambda_{min}\left\|\alpha^{k+1} - \alpha^k\right\|_2^2 \quad (22)\end{aligned}$$

**Proof of Theorem 2**

The strictly positive definiteness of $\bar{\mathbf{K}}$ guarantees $\lambda_{min} > 0$. (22) implies that $\{f(\alpha^k)\}$ is a decreasing sequence. Given that $f(\alpha^k) \geq -\frac{1}{2}\mathbf{y}^\top \bar{\mathbf{K}}^{-1}\mathbf{y} > -\infty$, we have that $\lim_{k\to\infty}(f(\alpha^k)) = -\frac{1}{2}\mathbf{y}^\top \bar{\mathbf{K}}^{-1}\mathbf{y}$. Applying (22) again, we obtain that $\{\alpha^{k+1} - \alpha^k\}$ converges to 0.

Since $f(\alpha)$ is a positive definite quadratic form, the set $\{\alpha|f(\alpha) \geq f(\alpha^0)\}$ is compact. $\{\alpha^k\}$ lies in this set, so it is a bounded sequence. Let $\bar{\alpha}$ be the limit of any convergent subsequence $\{\alpha^k\}$, $k \in \Gamma$. Since there are only a finite number of variables, there exists at least one active set $B^*$ which occurs infinitely in this subsequence. Let $\Gamma^* \subseteq \Gamma$ be the set of superscripts corresponding to $B^*$ and $s_1$ the index in $B^*$ first selected by GBCD. The $s_1$-th component of the gradient vector at $\bar{\alpha}$ is

$$\begin{aligned}g_{s_1}(\bar{\alpha}) &= \lim_{k\to\infty, k\in\Gamma^*} g_{s_1}\left(\alpha^k\right) \\ &= \lim_{k\to\infty, k\in\Gamma^*} g_{s_1}\left(\alpha^k\right) - g_{s_1}\left(\alpha^{k+1}\right) + \lim_{k\to\infty, k\in\Gamma^*} g_{s_1}\left(\alpha^{k+1}\right) \quad (23)\end{aligned}$$

Given $B(k)$, the active set at the $k$-th iteration, we have

$$\mathbf{g}_{B(k)}\left(\alpha^{k+1}\right) = \mathbf{g}_{B(k)}\left(\alpha^k\right) + \bar{\mathbf{K}}_{B(k)B(k)}\Delta\alpha_{B(k)}^{opt} \quad (24)$$

Substituting $\Delta\alpha_{B(k)}^{opt} = -\bar{\mathbf{K}}_{B(k)B(k)}^{-1}\mathbf{g}_{B(k)}^k$ into (24), we get $\mathbf{g}_{B(k)}(\alpha^{k+1}) = \mathbf{0}$. This indicates that $\lim_{k\to\infty, k\in\Gamma^*}(g_{s_1}(\alpha^{k+1})) = 0$. Since $\{\alpha^{k+1} - \alpha^k\}$ converges to 0, $\lim_{k\to\infty, k\in\Gamma^*}(g_{s_1}(\alpha^k) - g_{s_1}(\alpha^{k+1})) = 0$. Thus, we obtain

$$g_{s_1}(\bar{\alpha}) = 0 \quad (25)$$

From step 3 of the greedy approximation (we search exhaustively for the first active variable), we have

$$\frac{-g_{s_1}\left(\alpha^k\right)^2}{2\left(k(\mathbf{x}_{s_1}, \mathbf{x}_{s_1}) + \sigma^2\right)} \leq \frac{-g_i\left(\alpha^k\right)^2}{2\left(k(\mathbf{x}_i, \mathbf{x}_i) + \sigma^2\right)} \quad \forall i \in \{1, \cdots, n\} \quad (26)$$

Because $\bar{\mathbf{K}}$ is a positive definite matrix, $k(\mathbf{x}_i, \mathbf{x}_i) > 0$, $\forall i \in \{1, \cdots, n\}$. Taking the limit of (26), we get

$$\begin{aligned}g_i(\bar{\alpha})^2 &= \left[\lim_{k\to\infty, k\in\Gamma^*} g_i\left(\alpha^k\right)\right]^2 \\ &\leq \frac{k(\mathbf{x}_i, \mathbf{x}_i) + \sigma^2}{k(\mathbf{x}_{s_1}, \mathbf{x}_{s_1}) + \sigma^2}\left[\lim_{k\to\infty, k\in\Gamma^*} g_{s_1}\left(\alpha^k\right)\right]^2 \\ &= \frac{k(\mathbf{x}_i, \mathbf{x}_i) + \sigma^2}{k(\mathbf{x}_{s_1}, \mathbf{x}_{s_1}) + \sigma^2}\left[g_{s_1}(\bar{\alpha})\right]^2 \\ &= 0 \quad \forall i \in \{1, \cdots, n\} \quad (27)\end{aligned}$$

Thus, $\bar{\alpha}$ is the optimal solution of (3). Strict convexity of $f(\alpha)$ further implies that the sequence $\{\alpha^k\}$ itself converges to the global optimum of (3).

**Acknowledgements:** This work was supported, in part, by the EC and the NSF, under awards MCEXT-025481 and IIS-0535140.